\documentclass{ieeetran}
\usepackage{cite}
\usepackage{amsmath,amssymb,amsfonts}
\usepackage{algorithmic}
\usepackage{graphicx}
\usepackage{textcomp}
\usepackage{bm}
\makeatletter
\AtBeginDocument{\DeclareMathVersion{bold}
\SetSymbolFont{operators}{bold}{T1}{times}{b}{n}
\SetMathAlphabet{\mathrm}{bold}{T1}{times}{b}{n}
\SetMathAlphabet{\mathit}{bold}{T1}{times}{b}{it}
\SetMathAlphabet{\mathbf}{bold}{T1}{times}{b}{n}
\SetMathAlphabet{\mathtt}{bold}{OT1}{pcr}{b}{n}
\SetSymbolFont{symbols}{bold}{OMS}{cmsy}{b}{n}
\renewcommand\boldmath{\@nomath\boldmath\mathversion{bold}}}
\makeatother
\def\BibTeX{{\rm B\kern-.05em{\sc i\kern-.025em b}\kern-.08em
    T\kern-.1667em\lower.7ex\hbox{E}\kern-.125emX}}

\usepackage[utf8]{inputenc} 
\usepackage[T1]{fontenc}    
\usepackage{hyperref}       
\usepackage{url}            
\usepackage{amsfonts}       
\usepackage{nicefrac}       
\usepackage{microtype}      
\usepackage{xcolor}         

\usepackage{caption}
\usepackage{subcaption}
\usepackage{amsmath}        
\usepackage{graphicx}
\usepackage{pdfpages}

\usepackage{array}
\usepackage{booktabs}
\usepackage{multirow}
\usepackage{makecell}

\definecolor{customgreen}{RGB}{48,169,107}
\usepackage{pifont}

\newcommand{\greencheck}{\color{customgreen}{\ding{51}}}
\newcommand{\redx}{\color{red}{\ding{55}}}

\usepackage{enumitem}
\usepackage{lipsum}
\usepackage{comment}
\usepackage{flushend}

\begin{document}

\title{Datasheets for Machine Learning Sensors}

\author{Matthew Stewart$^{1,*}$, Yuke Zhang$^{2,*,\dagger}$, Pete Warden$^{3, 4}$, Yasmine Omri$^{3}$, Shvetank Prakash$^{1}$, Jacob Huckelberry$^{1}$,\\Joao Henrique Santos$^{1}$, Shawn Hymel$^{5}$, Benjamin Yeager Brown$^{1}$, Jim MacArthur$^{1}$, Nat Jeffries$^{4}$, Emanuel Moss$^{6}$, Mona Sloane$^{7}$, Brian Plancher$^{8,9}$, and Vijay Janapa Reddi$^{1}$%
\thanks{$^{1}$Harvard University, Cambridge, MA, USA}%
\thanks{$^{2}$University of Toronto, Toronto, ON, Canada}%
\thanks{$^{3}$Stanford University, Stanford, CA, USA}%
\thanks{$^{4}$Useful Sensors, USA}%
\thanks{$^{5}$Independent Researcher, USA}%
\thanks{$^{6}$Intel Labs, Hillsboro, OR, USA}%
\thanks{$^{7}$University of Virginia, Charlottesville, VA, USA}%
\thanks{$^{8}$Barnard College, Columbia University, New York, NY, USA}%
\thanks{$^{9}$Dartmouth College, Hanover, NH, USA}%
\thanks{$^{*}$These authors contributed equally to this work.}
\thanks{$^{\dagger}$ Corresponding author contact: \texttt{yuke.zhang@utoronto.ca}}}

\maketitle
\begin{abstract}
Machine learning (ML) is becoming prevalent in embedded AI sensing systems. These “ML sensors” enable context-sensitive, real-time data collection and decision-making across diverse applications ranging from anomaly detection in industrial settings to wildlife tracking for conservation efforts. As such, there is a need to provide transparency in the operation of such ML-enabled sensing systems through comprehensive documentation. This is needed to enable their reproducibility, to address new compliance and auditing regimes mandated in regulation and industry-specific policy, and to verify and validate the responsible nature of their operation. To address this gap, we introduce the datasheet for ML sensors framework. We provide a comprehensive template, collaboratively developed in academia–industry partnerships, that captures the distinct attributes of ML sensors, including hardware specifications, ML model and dataset characteristics, end-to-end performance metrics, and environmental impacts. Our framework addresses the continuous streaming nature of sensor data, real-time processing requirements, and embeds benchmarking methodologies that reflect real-world deployment conditions, ensuring practical viability. Aligned with the FAIR principles (Findability, Accessibility, Interoperability, and Reusability), our approach enhances the transparency and reusability of ML sensor documentation across academic, industrial, and regulatory domains. To show the application of our approach, we present two datasheets: the first for an open-source ML sensor designed in-house and the second for a commercial ML sensor developed by industry collaborators, both performing computer vision-based person detection.
\end{abstract}

\begin{figure}[!t]
     \centering
     \includegraphics[width=0.95\columnwidth]{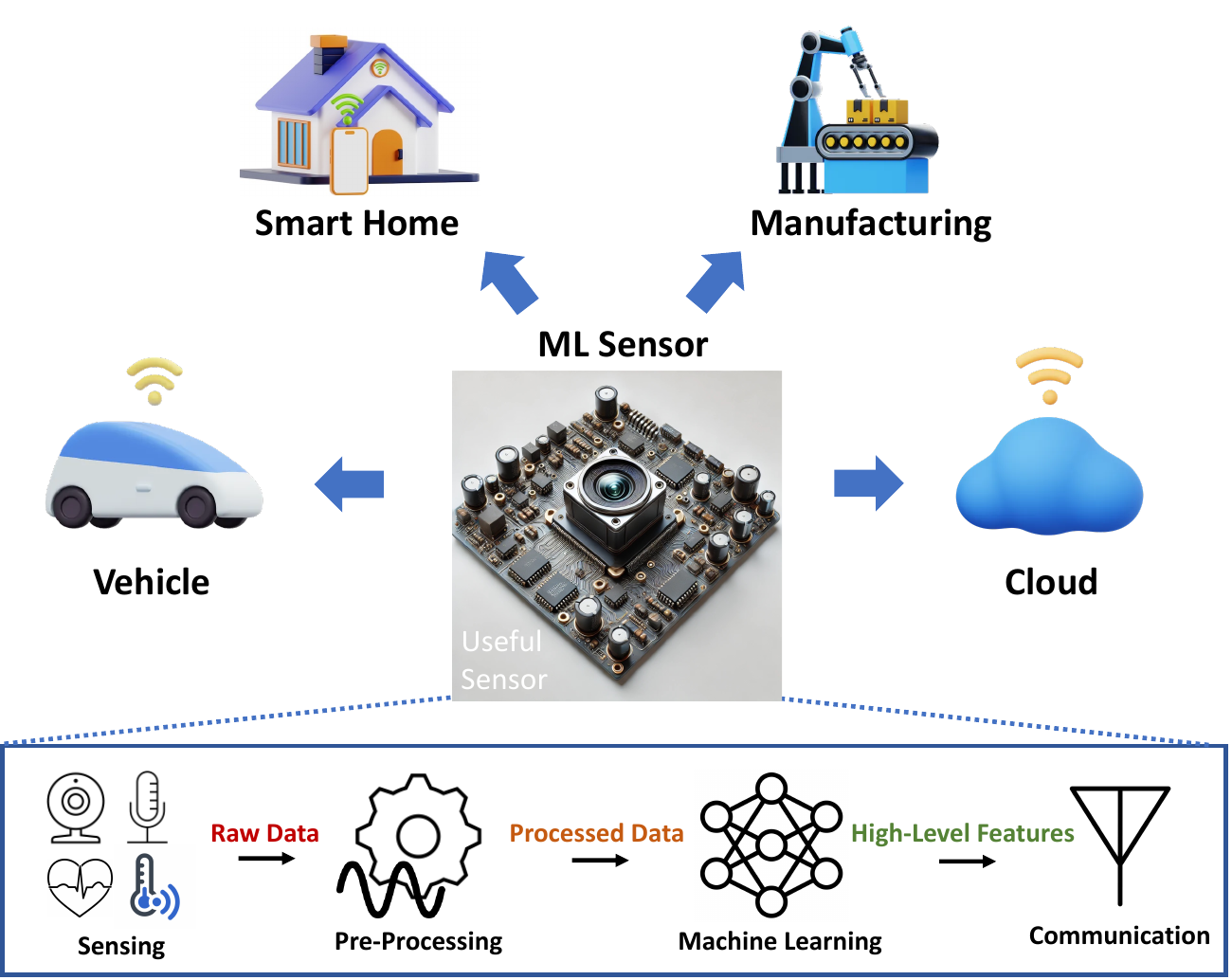}
     \caption{\footnotesize The ML sensor paradigm: deploying machine learning models directly on the sensor for privacy-promoting and energy-efficient edge intelligence.}
      \label{fig:mlsensor}
      \vspace{-20pt}
\end{figure}
\section{Introduction}
The merging of machine learning (ML), sophisticated data processing methods, and sensor technology is leading to rapid growth in the use of smart sensors~\cite{yadav2024advancements}, and the development of a new generation of intelligent sensing devices---ML sensors~\cite{warden23MLSensors}. An ML sensor is a standalone unit that employs ML algorithms directly on the device to handle intricate sensor data, providing real-time analysis and data-driven decision making at the point of data collection for complex tasks such as activity recognition~\cite{uddin2021human} and motion tracking~\cite{yang2022topographic}. Figure~\ref{fig:mlsensor} outlines its key components: sensing elements, pre-processing capabilities, ML model execution, and communication.
Importantly, unlike traditional sensors which require remote cloud processing, ML sensors deploy the ML model directly on the sensor, enabling both energy-efficient, and privacy-preserving, edge intelligence.

Although they represent an important advancement in gathering, analyzing, and responding to environmental data, there is no comprehensive framework for creating ML sensor-specific
transparency to support reproducibility, analysis of social impact, and accountability, e.g., ML audits~\cite{hartmann2024addressing, mokander2022conformity, falco2021governing}. However, such transparency can be achieved by building on existing documentation tools and approaches, serving as a bridge between ML sensor producers, end users, and academic researchers.
When ML sensor transparency takes the form of standardized documentation, academic researchers can better understand the practical needs and priorities of industry, aligning research with real-world applications, and supporting efforts to make ML sensors auditable and compliant with new and existing regulations (e.g., ML system audits increasingly required under the EU AI Act~\cite{european_commission_ai_2024,mokander2022conformity}).
Academic researchers can also better understand specifications and performance metrics to develop new applications, comparative studies, and further advancements in sensor technology.
At the same time, ML sensor producers can receive valuable feedback from users and researchers, informing future improvements and fostering innovation, while end-users can transparently understand the capabilities of the devices they purchase. 

There are existing documentation approaches that have the potential to provide high levels of transparency for ML sensors.
However, these existing frameworks often do not fully address the distinct complexities associated with ML sensors. For example, model cards~\cite{Mitchell_2019} tend to emphasize model design and performance metrics, but omit detailed discussions on data management and sensor privacy compliance. Similarly, traditional sensor datasheets for sensor hardware focus on technical details, such as power consumption and latency, without considering the ethical aspects of data collection and processing, or how sensors address bias and privacy issues.

\textbf{To bridge this gap, we introduce a ML sensor-specific datasheet framework that focuses on creating transparency for reproducibility and auditability of ML sensors.}
Importantly, our framework was developed by deploying a rigorous engagement approach focused on industry, academia, and government stakeholder engagement through seminars, workshops, and in-depth discussions, that was combined with an iterative refinement process through multiple rounds of testing.
The resulting ML sensor datasheet framework offers standardized sections that capture vital information about the sensor’s capabilities, data processing flows, and social impact dimensions.
To establish transparency for auditability, our datasheet offers comprehensive information about the ML sensor, covering every aspect from the sensing device and hardware specifications to the embedded ML model and system-level performance metrics.
Additionally, our datasheet introduces components on environmental impact and end-to-end performance, providing a holistic view of the sensor’s sustainability and operational resilience in diverse conditions. To enable public access, for transparency and feedback, we \emph{open-source} our datasheets at: \textcolor{blue}{\texttt{\href{https://github.com/harvard-edge/ML-Sensors}{github.com/harvard-edge/ML-Sensors}}}.

Our work supports standardizing the reporting of ML sensor features to facilitate transparency for auditability and to help developers and organizations follow best practices, support ongoing enhancements as standards progress, and ultimately encourage more responsible usage of ML sensors in sensitive and emergent contexts.
The contributions of our work are:
\begin{itemize}[]
    \item A novel  \textbf{ML sensor datasheet framework}, developed via industry, academia, and government stakeholder engagement and iterative refinement, that addresses the unique \textbf{transparency challenges of ML sensors} that are not fully captured by existing documentation approaches.
    
    \item Our \textbf{datasheet enables transparency for auditability} of ML sensors, bridging the gap between ML algorithms and embedded systems. It provides a \textbf{standardized approach} for documenting important aspects such as data management, privacy compliance, and social impacts, in addition to traditional technical specifications. 
    
    \item We demonstrate the \textbf{practical application of our framework} through two case studies: an \textbf{open-source ML sensor} and a \textbf{proprietary sensor}. These examples show how our datasheet can be effectively implemented across different types of ML sensors and establish \textbf{best practices} for future development and deployment of ML sensors.
\end{itemize}
\section{Background and Related Work} \label{sec:related}
\begin{table*}[!t]
\caption{Comparison of ML sensor datasheets with other datasheet types.}
\centering
\resizebox{\textwidth}{!}{
\begin{tabular}{lcccccc}
\toprule
Datasheets $\downarrow$ & Hardware & Privacy \& Security & Dataset & Model & Env. Impact & End-to-End \\\midrule
ML Sensors (our work) 
    & \greencheck & \greencheck & \greencheck & \greencheck & \greencheck & \greencheck \\\midrule
Traditional Sensor Datasheet 
    & \greencheck & \redx & \redx & \redx & \redx & \redx \\
IoT Security/Privacy Label \cite{emami2020ask,emami2021informative}
    & \redx & \greencheck & \redx & \redx & \redx & \redx \\
Data Nutrition Label \cite{holland2018dataset,chmielinski2022dataset}
    & \redx & \redx & \greencheck & \redx & \redx & \redx \\
Datasheets for Datasets \cite{gebru2021datasheets}
    & \redx & \redx & \greencheck & \redx & \redx & \redx \\ 
Model Cards \cite{Mitchell_2019}
    & \redx & \redx & \redx & \greencheck & \redx & \redx \\
\bottomrule
\end{tabular}
}
\label{table:related}
\end{table*}

\subsection{Machine Learning Sensors} 

Embedded sensing systems have a long history and serve as fundamental building blocks for many applications~\cite{warden23MLSensors,prakash2023tinyml}, converting physical phenomena into signals for processing. 
Embedded AI sensing has marked a significant evolution, introducing sophisticated on-device processing capabilities that can extract meaningful insights directly from sensor data.
However, this evolution also introduced new intertwined challenges of hardware optimization, ML deployment, and system design that developers must balance~\cite{warden23MLSensors}, creating significant barriers to entry.

\begin{figure}[!t]
    \centering
    \includegraphics[width=0.9\columnwidth]{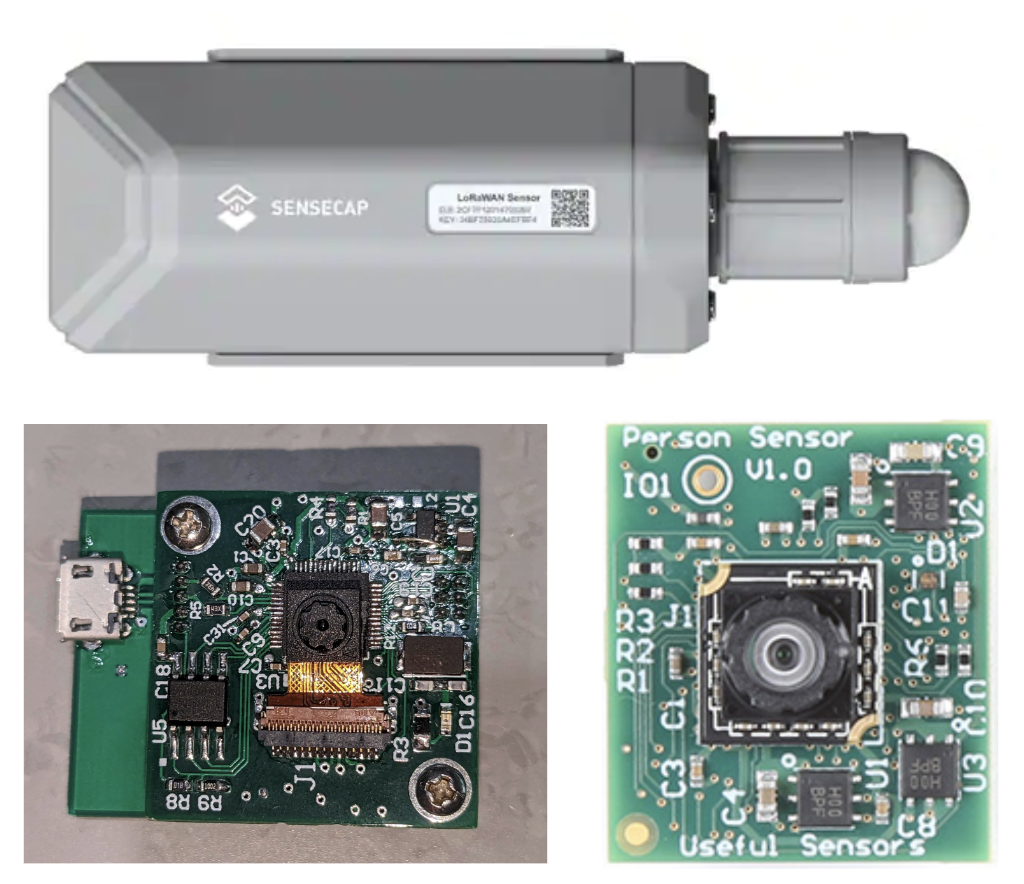}
    \caption{\footnotesize Examples of existing ML sensors; \textit{(top)} Seeed Studio's SenseCAP LoRaWAN sensor~\cite{seeed_panda_detector} for long-range data collection IoT scenarios like smart farming, \textit{(bottom left)} our own person detection sensor whose design is open-source (link redacted for double blind), and \textit{(bottom right)}, Useful Sensor's person sensor~\cite{useful_sensors1}. We use these person detection sensors for our case study applying our ML sensor datasheet to real ML sensors in Section~\ref{sec:caseStudy}.}
    \label{fig:personsensor}
     \vspace{-10pt}
\end{figure}

The ``ML sensors'' paradigm reimagines this evolution by elevating the abstraction level of embedded AI development~\cite{warden23MLSensors, sloane2025materiality}. 
As illustrated in Figure~\ref{fig:personsensor}, existing commercial ML sensors encapsulate both sensing and computation capabilities within a unified package. These specialized devices focus exclusively on performing ML inference on physical phenomena, communicating only processed outcomes through a streamlined interface to the broader system. 

This change transforms embedded AI development through several innovations. The unified design improves the locality of the data and the processing efficiency while significantly improving the modularity and maintainability of the system. The encapsulated approach strengthens privacy and security by limiting the exposure of raw sensor data, while the streamlined interface simplifies reasoning about data flows and system behavior. This modular and self-contained approach to embedded AI significantly simplifies the documentation of technical specifications, model characteristics, and operational constraints through our proposed datasheets as it enables standalone evaluation of the ML sensor product absent from its interaction with large-scale cloud-based data processing. 

As such, just as engineers today can readily integrate standard temperature or pressure sensors into their designs, with ML sensors, \emph{that have transparent and comprehensive datasheets}, we envision a future where developers can similarly incorporate pre-packaged ``person sensors'' or ``gesture sensor.''

\subsection{Datasheets and Documentation in AI}

Traditional ML benchmarks and datasheets have predominantly focused on software models, covering data collection, cleaning, labeling, and intended use, both for specific datasets~\cite{bandy2021addressing, zilka2022survey, srinivasan2021artsheets}, and for more general frameworks for documenting dataset characteristics~\cite{gebru2021datasheets,holland2018dataset,chmielinski2022dataset}. While efforts have been made to document various features of ML services~\cite{arnold2019factsheets}, provide benchmarked evaluations for ML models~\cite{Mitchell_2019}, and include relevant privacy and security information for IoT devices~\cite{emami2020ask,emami2021informative}, these approaches fall short in capturing the unique challenges posed by the integration of hardware and software in ML sensors. Moreover, recent efforts to 
highlight potential social impacts of ML and sensor devices~\cite{boyd2021datasheets,owen2013framework,sloane2025materiality}
as well as efforts to characterize operational and embodied emissions of hardware devices~\cite{gupta2022act,prakash2023tinyml}, further underscores the need for a new comprehensive datasheet format. 
As shown in Table~\ref{table:related}, ML sensors uniquely encompass integrated hardware, software, and machine learning elements, necessitating a novel approach that amalgamates diverse concepts. Therefore, build on prior work and augment traditional sensor datasheets with vital information relevant for ML and IoT, as well as other new elements, resulting in a comprehensive and extendable datasheet that can effectively address the challenges of transparency and auditability in this emerging paradigm.
\section{Datasheet Design}
\label{sec:datasheet}
\subsection{Design Motivation and Methodology}
ML-enabled sensors and devices often provide datasheets with hardware specifications and instructions for using the sensor capabilities. However, these existing datasheets often lack substantial details regarding: (1) information on data characteristics, distributions, and potential biases which would build trust in model performance; (2) model architecture and benchmarks, with details on design choices and performance evaluations on standard datasets allowing better assessment of expected capabilities; (3) environmental impact over the device lifetime, including materials, energy use, and recyclability allowing more sustainable adoption; (4) robustness to anticipated changes in operating conditions once deployed, with testing results on transformations in factors like lighting, backgrounds, weather, and wear guiding appropriate scenarios; and (5) aspects of privacy and security, such as disclosure and protections around data sharing, vulnerabilities, and compliance.

Creating transparency by providing documentation can increase user trust, facilitate accountability, and encourage responsible development and adoption of ML innovations across industries~\cite{Mitchell_2019}. It also enables those implementing ML systems to  ensure that outcomes align with their principles, policies, compliance responsibilities. The guiding motivation is to clearly communicate the inner workings of the ML sensor, including details about its embedded ML models, training data, and performance metrics to ensure that users, developers,  stakeholders, and auditors can easily understand how the sensor processes data, makes decisions, and performs under various conditions. To support this, the datasheet includes descriptions of AI models and end-to-end performance analysis summaries that outline how the sensor operates in different environments, as well as sections that document bias mitigation strategies, privacy safeguards, and compliance standards and policies. 

In line with the FAIR principles, we ensure that ML sensor datasheets are structured to promote discoverability, integration, and reuse across diverse contexts. These principles are important as ML sensors become part of broader ecosystems that require standardization, transparency, and traceability. Therefore, the development of our datasheet framework for ML sensors was based on a rigorous and multifaceted \textit{engagement process with industry, academia, and government stakeholders} and \textit{iterative refinement}. This process ensured a holistic approach the design process in order to address the complex challenges of ML sensor documentation.

\textbf{Stakeholder Engagement.} We engaged with a broad spectrum of stakeholders, ranging from multinational corporations (e.g., Google, Dyson, LG, Samsung), to emerging startups, to government entities, to academia. The diverse industry and government participation was crucial in capturing real-world implementation needs on various operational scales and use cases. Our academic engagement centered on a two-day seminar
 at the Harvard Radcliffe Institute, titled "Safeguarding User Privacy in the era of Sensor Intelligence". 
This event brought together experts from multiple disciplines to critically examine risks and power structures associated with ML sensors, informing the essential components of our data sheet. Finally, to ensure compliance with international legal standards and regulations, we incorporated information on data protection and privacy regulations into our framework (e.g., GDPR~\cite{GDPR2016}).

\textbf{Iterative Refinement.} The overall development of our ML sensor datasheet framework was not a linear process, but rather an iterative process. Our initial template, while comprehensive, served as a starting point for a series of refinements that would shape it into a more robust and universally applicable tool.
Throughout this refinement process, we leveraged multiple channels of feedback and expertise.

Our ML sensor-focused workshops held at the TinyML Summit~\cite{TinyMLsummit2023} have proved to be useful forums for engaging with a diverse community of practitioners and researchers. These workshops provided a platform for lively discussions and critical evaluations of our framework, offering insights from various perspectives and use cases that we might not have initially considered.
Parallel to these workshops, we entered into a series of in-depth discussions with the National Institute of Standards and Technology (NIST)~\cite{nist}. These conversations helped us align our framework with emerging standards and best practices in the field of ML and sensor technologies. The expertise provided by NIST was instrumental in ensuring that our datasheet not only met current needs but was also positioned to adapt to future developments in the rapidly evolving landscape.

The most crucial aspect of our refinement process was the practical application of the datasheet to real-world ML sensor products. We tested our framework on a range of devices, from open-source projects to commercial products, spanning various applications and complexities. This hands-on approach, discussed in greater detail in the case studies (Section~\ref{sec:caseStudy_ai_components}), allowed us to identify gaps in our documentation, uncover unforeseen challenges, and validate the practical utility of our framework in different contexts.

Each iteration of testing and feedback led to adjustments in our datasheet structure, content, and guidance. We fine-tuned categories, clarified definitions, and added new sections to address emerging concerns. This iterative process of application, feedback, and refinement was crucial to transform our initial concept into a comprehensive, flexible, and practical tool to document ML sensors.

\textbf{Key Learnings.} Our learnings from academia include: (a) the importance of addressing bias in sensor data collection; (b) the need for transparency in algorithmic decision-making processes; (c) emphasis on the sustainability and ethical responsibility.  Our insights from industry and government include: (a) the importance of real-time capabilities in ML sensors; (b) the need for clear and transparent metrics; (c) emphasis on integration with existing sensor ecosystems, regulations, and frameworks.

\textbf{Accessibility and Maintenance.} We have open-sourced our datasheet to enhance maintenance, distribution, and accessibility. All stakeholders have easy access to the most current version of our datasheet. We gather feedback from all relevant stakeholders and update the datasheet to reflect any changes.

\textbf{Outcome.} The result of this process is a datasheet framework that is stakeholder-driven and practically validated across a spectrum of real-world applications. However, we recognize that the ML sensors is dynamic and ever-evolving. As such, we view our current framework not as a final product, but as a robust foundation ready for further adaptation and improvement as the technology and its applications continue to advance.

\subsection{Overview of the Proposed Datasheet}

\begin{figure}[!t]
    \centering
    \includegraphics[width=0.9\linewidth]{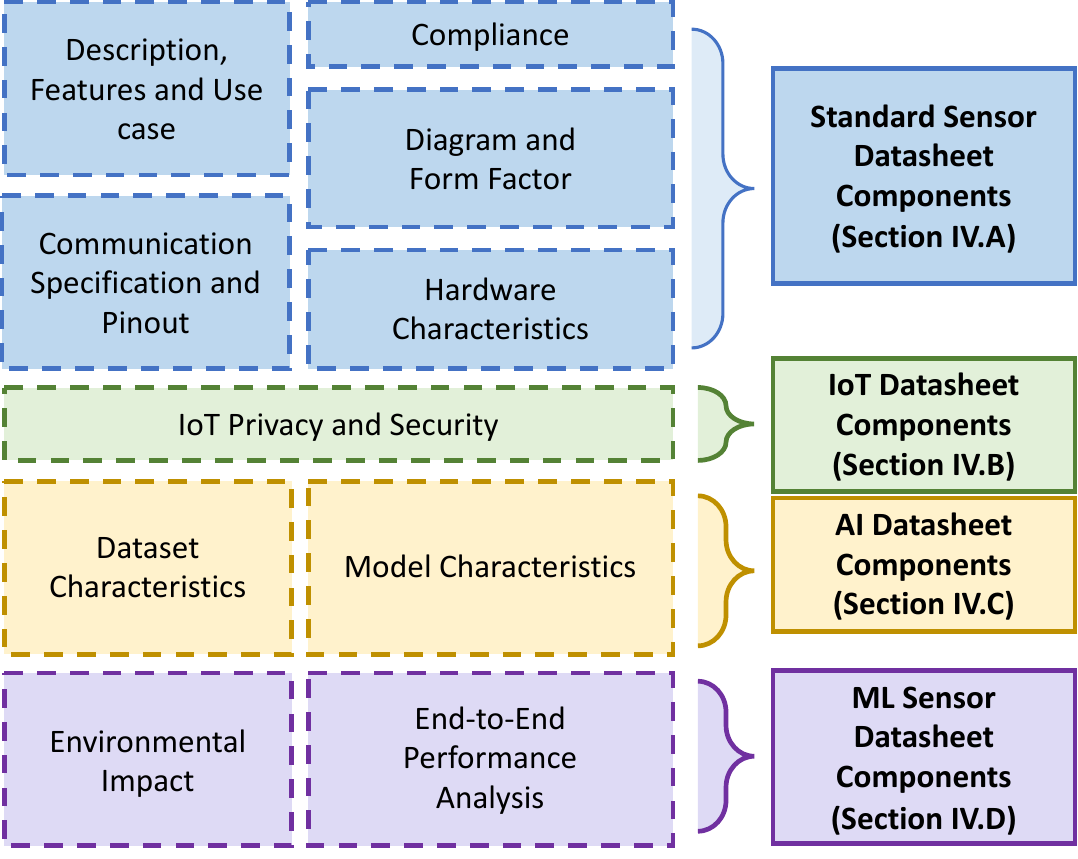}
    \caption{\footnotesize Schematic of the proposed datasheet template for ML sensors.}
    \label{fig:datasheet}
    \vspace{-10pt}
\end{figure}
Based on the principles and design methodology, our datasheet, shown in Figure~\ref{fig:datasheet}, is segmented into these areas:
\begin{itemize}[noitemsep, topsep=0pt]
    \item \textbf{Standard Sensor Datasheet Components.} This section mirrors conventional electronics datasheets, encapsulating core sensor information such as a detailed description, features, use cases, communication specifications, pinout details (the function of each connector or pin), compliance with industry standards, as well as physical attributes like diagrams, form factors, and hardware characteristics.
    \item \textbf{IoT Datasheet Components.} Dedicated to the integration of the sensor within Internet of Things ecosystems, this part outlines privacy and security protocols specific to IoT, ensuring that the sensor's deployment aligns with modern cybersecurity practices.
    \item \textbf{AI Datasheet Components.} This segment provides in-depth information about the AI models embedded in the sensor, including the type of algorithms used, the training data, and any pertinent model characteristics that users and developers should be aware of.
    \item \textbf{ML Sensor Datasheet Components.} To address the needs of ML sensors, this section adds more layers of analysis, emphasizing documentation of the ML model as deployed on the specific sensor, which makes the environmental and end-to-end performance analysis possible. Our datasheet
    includes the environmental impact, showing the device's footprint and sustainability, and an end-to-end performance analysis which reviews the sensor's functionality in real-world scenarios, considering factors like varying environmental conditions and fairness parameters.
\end{itemize}

Different from prior work~\cite{Mitchell_2019, gebru2021datasheets, emami2020ask, emami2021informative, chmielinski2022dataset, holland2018dataset}, our datasheet presents hardware characteristics, details about the embedded ML model, IoT privacy and security protocols, and, notably, the system-level ML sensor features of environmental impact and end-to-end performance analysis. These system-level features were included based on feedback from seminars and workshops, emphasizing the increasingly urgent need for sustainable design and reliable performance metrics in real-world applications. Our datasheet format also aligns with the FAIR principles by making sensor metadata findable and well-documented, enabling accessible and machine-readable formats, promoting interoperability with existing documentation standards, and supporting the reuse of sensor components, datasets, and models in diverse deployments. 
In Sections~\ref{sec:standard_components}-\ref{sec:mlsensor_components}, we provide a detailed breakdown of the four key components.
\section{Datasheet Implementation}
In this section, we provide a detailed breakdown of the four key components of the proposed datasheet and clarify the distinct elements relevant to the nine targeted tasks.


\subsection{Standard Datasheet Components} \label{sec:standard_components}

Standard datasheets provide general electronic and physical specifications to help developers, users, and auditors understand the device's basic capabilities, requirements, and form factor (i.e., its size, shape and layout).

\textbf{Description, Features, and Use Case.}
\textit{What are high-level characteristics of the sensor?} 
The description section of the ML sensor datasheet provides an introduction to the device for both technical and non-technical audiences. On the technical side, it includes details about the device's specifications, architecture, and operational principles. For non-technical readers, it offers a more accessible description, explaining the sensor's purpose and function in plain language. This section also highlights key features of the  device and a list of common applications of the sensor.

\textbf{Diagrams and Communication Specification.}
\textit{What does the device size, shape, and layout look like?} 
This section provides visual depictions and physical dimensions of the device. It includes detailed diagrams that illustrate the sensor's internal components and their interconnections, offering insights into the design and operation of the sensor. For non-technical audiences, these diagrams can provide a more intuitive understanding of the device, beyond what text descriptions can offer. These diagrams include form factor information which describes the physical shape, size, and layout of the sensor. This data is crucial in planning the sensor's integration into various systems and devices. 

\textbf{Hardware Characteristics.}
\textit{What components are a part of the device? How do they operate?} 
This section provides an overview of the physical and functional attributes of the device. It contains specifics about the sensor's integral hardware components, including the processor type, memory capacity, power requirements, and durability under different environmental conditions. In addition, it includes detailed information about the communication protocols supported by the sensor, such as Wi-Fi, Bluetooth, or cellular connectivity, along with data transfer rates. This data is crucial in determining the sensor's compatibility with existing hardware infrastructure and intended deployment environments.

\textbf{Compliance and Certification.} \label{sec:regulation}
\textit{Which international regulations and industry standards does the device conform to?} 
This section lists the certifications the sensor has achieved, signifying thorough testing and validation by recognized certification bodies. Compliance with these standards vouches for the sensor's reliability, safety, and overall quality, instilling confidence in developers and end-users about its dependable operation. These may encompass international data privacy regulations like GDPR~\cite{parliament2016regulation}, radio frequency usage guidelines like FCC regulations, or industry-specific requirements like HIPAA~\cite{act1996health} or FDA standards in healthcare. This section could also document the ML sensor's adherence to voluntary industry-specific best practices, such as ISO 26262 standard \cite{iso201826262} for autonomous vehicles. Additionally, given the rapidly evolving regulatory landscape for AI-driven technologies, frameworks like the EU AI Act are increasingly relevant. Anticipating such developments and providing transparency via the datasheet ensures the sensor and any systems the sensor is embedded in can be audited for compliance  with forthcoming standards and facilitates comprehensive, future-proof documentation. 


\subsection{IoT Datasheet Components} \label{sec:iot_components}

This section summarizes the ML sensor's safeguards and risks from an IoT perspective, facilitating awareness and informed decision-making when purchasing smart devices.

\textbf{Security and Privacy.}
\textit{What IoT security and privacy features does the ML sensor have?} This section promotes transparency and empower consumers, allowing them to make well-informed choices in an increasingly connected world. This section is structured in two distinct layers: a primary layer, which conveys essential privacy and security information in a concise and easily digestible manner, and a secondary layer, which delves into further detail for experts and more technically inclined users~\cite{emami2021informative}. 
Combined, the layers cover privacy-related aspects such as data collection, retention and transmission practices, security mechanisms (e.g., encryption, automatic security updates), as well as the types of sensors present on the device and their associated data modalities.
Information regarding the ability to update the device's ML models could also be included in this section, including details about the frequency, method, and security measures of these updates can further inform users about the longevity and reliability of the device's performance.

In addition to the primary privacy and security elements, other crucial issues may include the risk of unauthorized data access~\cite{aljabri2023machine}, vulnerabilities related to device authentication~\cite{de2013back} and cyberattacks~\cite{talaei2022dynamic, huckelberry2024tinyml}.
The datasheet can incorporate information regarding measures taken to mitigate these threats, such as multi-factor authentication protocols and tamper detection mechanisms.
Furthermore, in privacy-sensitive domains like healthcare or surveillance, 
datasheets should include results from privacy impact assessments and audits, which demonstrate the sensor's compliance with stringent data protection frameworks~\cite{zhang2023sal}. This expanded view ensures that consumers and technical stakeholders have a comprehensive understanding of how the ML sensor aligns with best practices for safeguarding sensitive data in specific use cases.


\subsection{AI Datasheet Components} \label{ai_components}

This section details the data used for training and the resulting model, enabling stakeholders to evaluate aspects like sampling, measurement, and bias.

\textbf{Dataset Characteristics.}
\textit{What data is the model trained on?} 
Outlining dataset characteristics is fundamental for ensuring transparency in ML systems. This transparency is crucial because it allows users, developers, and regulators to understand how and why a ML model makes certain decisions, and to assess its fairness, accuracy, and potential biases. By clearly detailing the nature of the training data, stakeholders can better evaluate the model's applicability to real-world scenarios and audit the dataset for how well it corresponds with its context of use. Several approaches have been proposed to achieve this, including datasheets for datasets~\cite{gebru2021datasheets} and the data nutrition label~\cite{holland2018dataset}.

Taking the data nutrition label as an example, this label communicates high-level dataset information to end-users, including (1) the sources of the dataset (i.e., governmental, commercial, academic), (2) licensing details of the dataset, (3) data modality, and (4) context-specific information (e.g., human-labeled, contains information about human individuals), amongst other information. 
This provides information about the context, content, and quality of dataset(s) used in training the ML model. As such, it makes it easier for developers, researchers, and stakeholders to assess data quality and potential biases such as sampling, measurement, and label bias~\cite{jiang2020identifying}.

\textbf{Model Characteristics.}
\textit{What are the characteristics of the trained model? What are the training method and hyperparameters?} This section of the datasheet provides insights into the specific ML model operating within the sensor. This includes important details such as the type of the ML model used, the size of the model in terms of parameters, the type and size of input data it can process, and the nature of output it generates. This section also discusses the model's performance metrics, such as accuracy, precision, recall, F1 score, and performance across demographic categories, measured on a relevant validation dataset.
It may also address the model's robustness to variations in input data, its sensitivity to noise, and its generalization capabilities. When applicable and not confidential, details regarding the training parameters used—such as learning rates, batch sizes, and training duration—can be included to provide a more complete understanding of the model’s development. This section is vital for users to understand the underlying ML model and its suitability for their specific use cases.


\subsection{ML Sensor Datasheet Components} \label{sec:mlsensor_components}

As IoT proliferation increases e-waste, consumers must be informed of devices' environmental sustainability. Similarly, as these devices are deployed in increasingly diverse environments, users must be able to understand how their device's end-to-end performance will vary. As such, this section outlines the envrionmental impact and end-to-end performance of deploying these devices across real-world settings, enabling the full embodied device to be evaluated by end-users.

\textbf{Environmental Impact.}
\textit{How does the device affect the environment during its lifetime?}
The Environmental impact component in the ML sensor datasheet provides insights into the sustainability and ecological footprint of ML sensors across different tasks. As ML sensors become widely deployed in various industries, transparency into their environmental impact is crucial for responsible design and usage.

The key element for environmental impact is \textit{Emissions and Carbon Footprint}.
To calculate the emissions and carbon footprint of an ML sensor, it’s essential to assess its environmental impact throughout its lifecycle—from production and usage to disposal. This begins with gathering the data on material sourcing and manufacturing processes, identifying the materials used (e.g., metals, plastics, semiconductors) and obtaining emissions data associated with their extraction, processing, and assembly. Life cycle assessment (LCA) simulators like the OpenLCA~\cite{OpenLCA} and the TinyML footprint calculator~\cite{prakash2023tinyml} can be used to compute comprehensive modeling of a product's lifecycle emissions.

Other elements within the environmental impact section of the datasheet can include \textit{materials and resource use} and \textit{end-of-life disposal}. 
\textit{Materials and resource use} list the materials used in each sensor’s construction, highlighting any eco-friendly or sustainably sourced components, which is especially important for biometric and environmental sensors in healthcare and monitoring applications that operate in sensitive environments. 
\textit{End-of-life disposal} instructions specify disposal instructions, recyclability of components, and any options for refurbishment or recycling, helping minimize environmental impact after the sensor reaches end-of-life.

\textbf{End-to-End Performance Analysis.}
\textit{How does the device perform as a whole with changing environmental parameters?} This section provides an encompassing evaluation of the sensor's performance from data acquisition to data processing and output generation. This holistic performance analysis may include metrics such as data collection rate, latency, power consumption, and accuracy of the sensor's outputs under a range of conditions. Additionally, it highlights the performance of the ML model when deployed on the sensor hardware, taking into account aspects such as data preprocessing, inference speed, and model accuracy. The analysis could also encompass how the sensor's performance scales with changes in workload or environmental conditions. 

\begin{figure*}[t!]
    \centering
    \includegraphics[width=\textwidth]{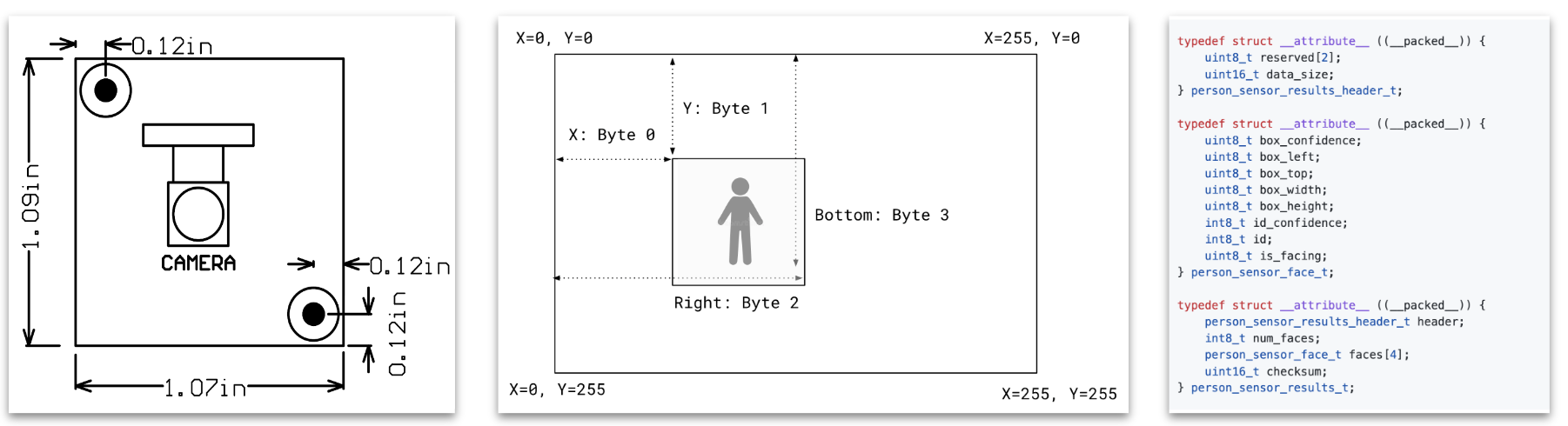}
    \caption{\footnotesize \textit{(left)} Device diagram of person detection ML sensor, \textit{(middle)} standard for data communication, and \textit{(right)} schema for communication of data off-sensor.}
    \label{fig:diagrams}
    \vspace{-5pt}
\end{figure*}

This section is important as it helps potential users understand not only the isolated performance of the sensor's components but also how they work together to provide a coherent service. This understanding is vital when integrating the sensor into larger systems or evaluating its fit for particular use-cases. 
Common factors in end-to-end performance analysis for various ML sensor tasks are 
demographic bias, distance, class diversity, and environmental conditions.

Performance metrics under these conditions include ML performance metrics such as classification accuracy, detection precision, and recognition accuracy, as well as hardware-specific measurements like power consumption, latency, and throughput. Depending on the intended application, ML sensor manufacturers may adapt these metrics to provide a customized end-to-end performance analysis.
This structured approach, combined with adherence to established standards, ensures that the end-to-end performance analysis provides meaningful insights into the sensor’s real-world reliability and efficiency.
\section{A Case Study in Person Detection} \label{sec:caseStudy}

In this section, we present a detailed case study of two person detection ML sensors (shown previously in Figure~\ref{fig:personsensor}), which are used in applications such as smart farming and occupancy monitoring. These sensors detect the presence of a person without identifying individuals or performing facial recognition. The first is an open-source ML sensor, developed as part of this research initiative.
The second is an ML sensor developed by a commercial partner.
This partnership allowed us to integrate academic research insights with practical, industry-standard design approaches. The result is a sensor that not only has commercial viability but also achieves increased auditability and transparency. 
The juxtaposition of these two sensors serves both provides a comprehensive overview of the current state of person detection technology, as well as offers insights into the different design and development methodologies employed in academic versus commercial settings. Finally, this case study aims to highlight the potential synergies between academic research and industry practices, suggesting a model for future collaborative efforts in the fields of ML and sensor technology. We \emph{open-source} the resulting datasheets at: \textcolor{blue}{\texttt{\href{https://github.com/harvard-edge/ML-Sensors}{github.com/harvard-edge/ML-Sensors}}}


\subsection{Standard Datasheet Components} \label{sec:caseStudy_standard_components}

In the context of our person detection sensor (Figure~\ref{fig:personsensor}), the description would be ``a device that predicts whether an individual is present in the view of the camera and outputs a corresponding signal response.'' Examples of diagrams and hardware specifications for our open-source person detector are shown in Figure~\ref{fig:diagrams}.
Figure~\ref{fig:diagrams} \textit{(left)} shows our ML sensor with a square form factor and dimensions $27.2 \textrm{mm} \times \textrm{27.7 mm}$. It employs the industry-standard Inter-Integrated Circuit ($\textrm{I}^2 \textrm{C}$) interface via a Qwiic connector~\cite{QwiicCon49:online}, allowing a data transfer rate of up to 100 kB/s. Figure~\ref{fig:diagrams} \textit{(middle)} and \textit{(right)} show the data standard and open-source schema we developed for communication~\cite{GitHubus32:online}. The sensor communicates through a single byte with values from 0 to 255. The device can accept power at 3.5-5.5 Vs with a 40 mA operating current. 

Additionally, while our own ML sensor has not obtained specific certifications or verification, the commercial sensors complies with RoHS~\cite{RoHS} for environmental safety and GDPR~\cite{GDPR} for data privacy and has been audited by Kodelski Security~\cite{kudelskisecurity} for security and privacy implications.

\subsection{IoT Datasheet Components} \label{sec:caseStudy_iot_components}
\begin{figure*}[!t]
    \centering
    \includegraphics[width=0.4\textwidth]{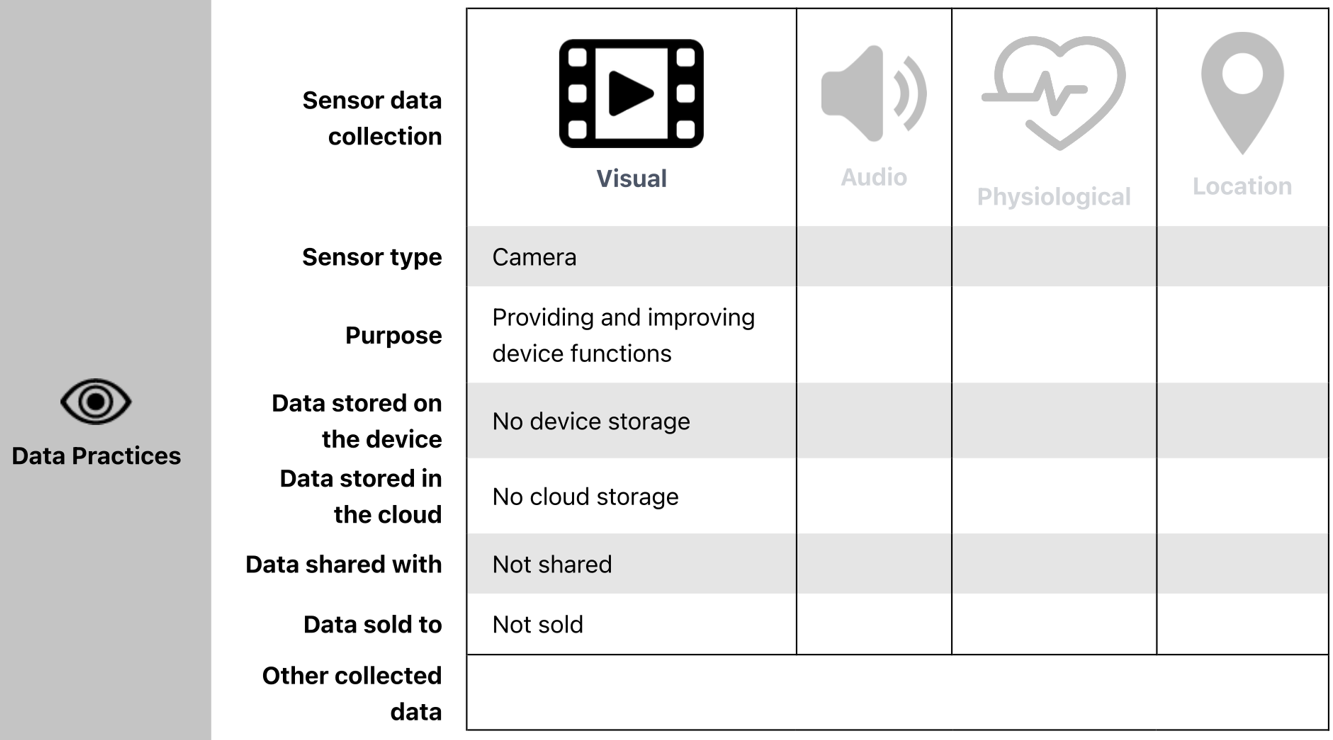}
    \hspace{0.1cm}
    \includegraphics[width=0.21\textwidth]{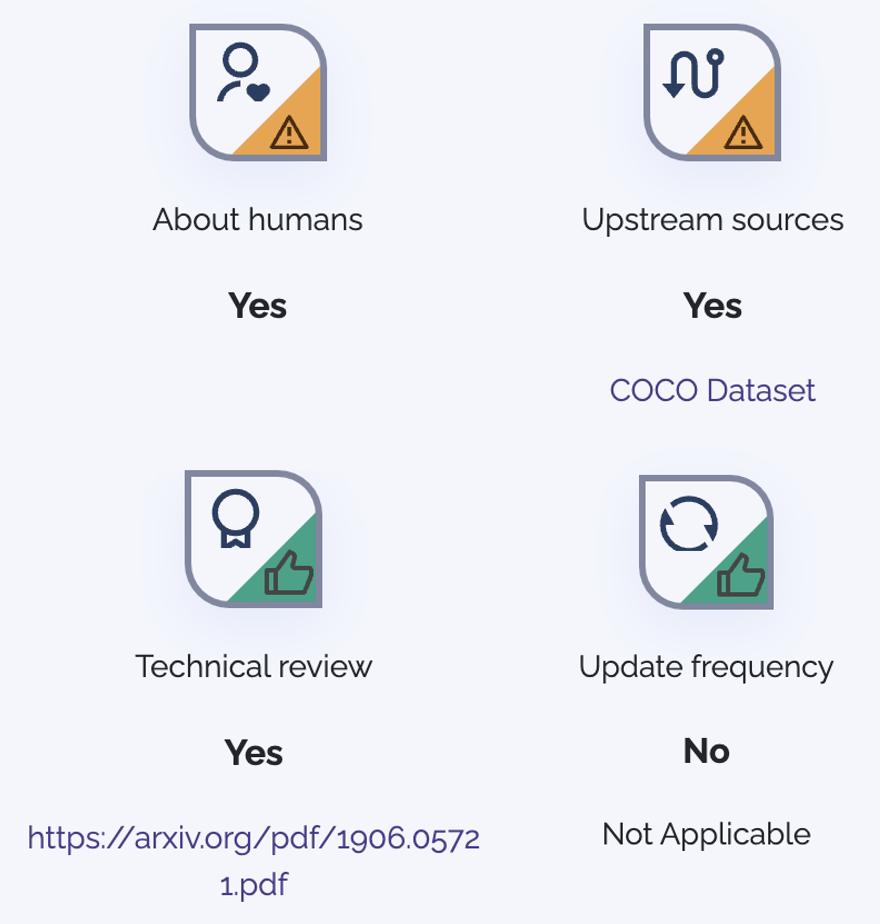}
    \hspace{0.1cm}
    \includegraphics[width=0.33\textwidth]{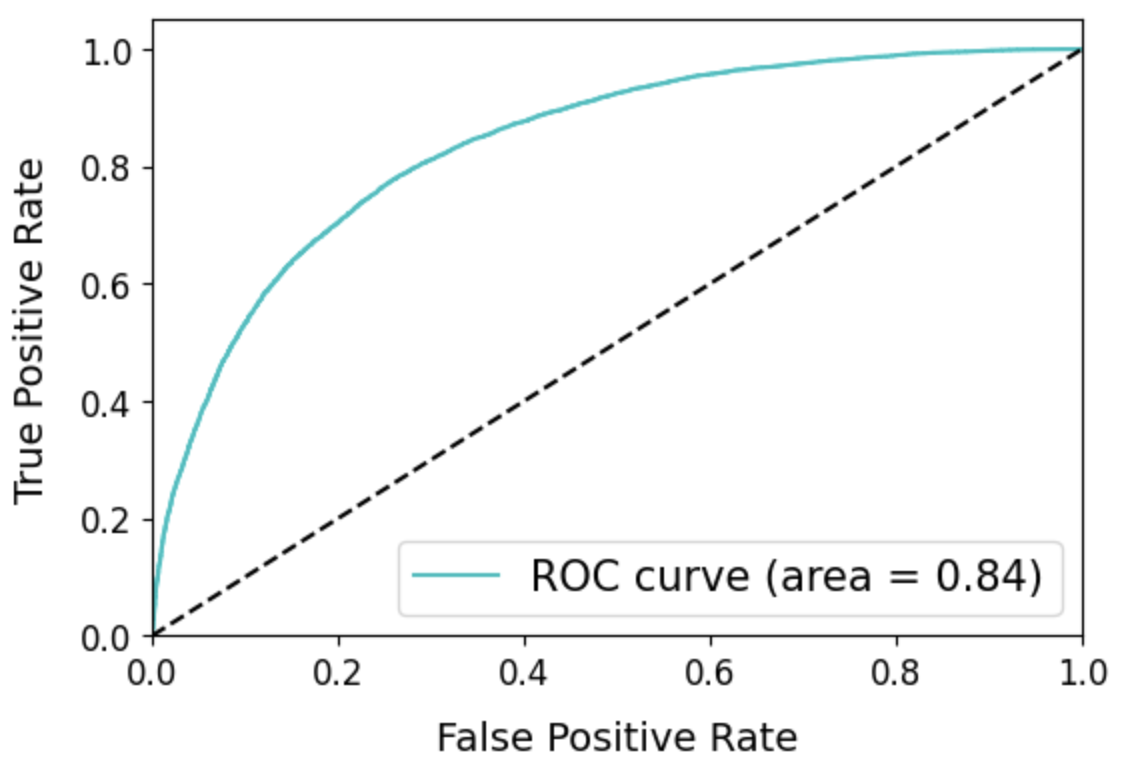}
    \caption{\footnotesize Primary IoT security and privacy label for the open-source person detection ML sensor \textit{(left)}, as well as its data nutrition label summary statistics \textit{(center)}, and the ROC curve of the person detection model evaluated on a test set \textit{(right)}.}
    \label{fig:security_label}
\end{figure*}

For our ML sensor, the IoT security and privacy label (see \autoref{fig:security_label} left) shows that there is only a camera on the device collecting data continuously, but this data is not being stored or transmitted off-device. The self-contained nature of the ML sensors means that they have limited networking capabilities, and thus privacy concerns from the transmission of raw data are minimal.
We note that the commercial sensor has an equivalent IoT privacy and security label.

\subsection{AI Datasheet Components} \label{sec:caseStudy_ai_components}
To evaluate the dataset used in training the on-device model, we utilize the second-generation Dataset Nutrition Label~\cite{holland2018dataset, chmielinski2022dataset}. Summary statistics for the data nutrition label for the open-source person detection sensor are shown on the right side of Figure \ref{fig:security_label}, and the full label is available in our open-sourced datasheet. This label highlights that the dataset, Visual Wake Words~\cite{chowdhery2019visual}, is from an upstream source (MS-COCO~\cite{lin2015microsoft}), contains information about humans obtained without consent, and that the dataset is not currently managed or updated.
Figure \ref{fig:security_label} also shows an example model characteristic of the ML sensor running a MobileNetV1 architecture \cite{howard2017mobilenets} trained for person detection. In particular, the ROC curve shows that the optimal threshold value lies around 0.52 to balance false positives and negatives.
The commercial sensor has a similar, but more complex, software architecture, resulting in a 91.8\% accuracy with a threshold of 0.7.

\subsection{ML Sensor Datasheet Components} \label{sec:caseStudy_mlsensor_components}

\begin{figure*}[h!]
    \centering
    \includegraphics[width=\textwidth]{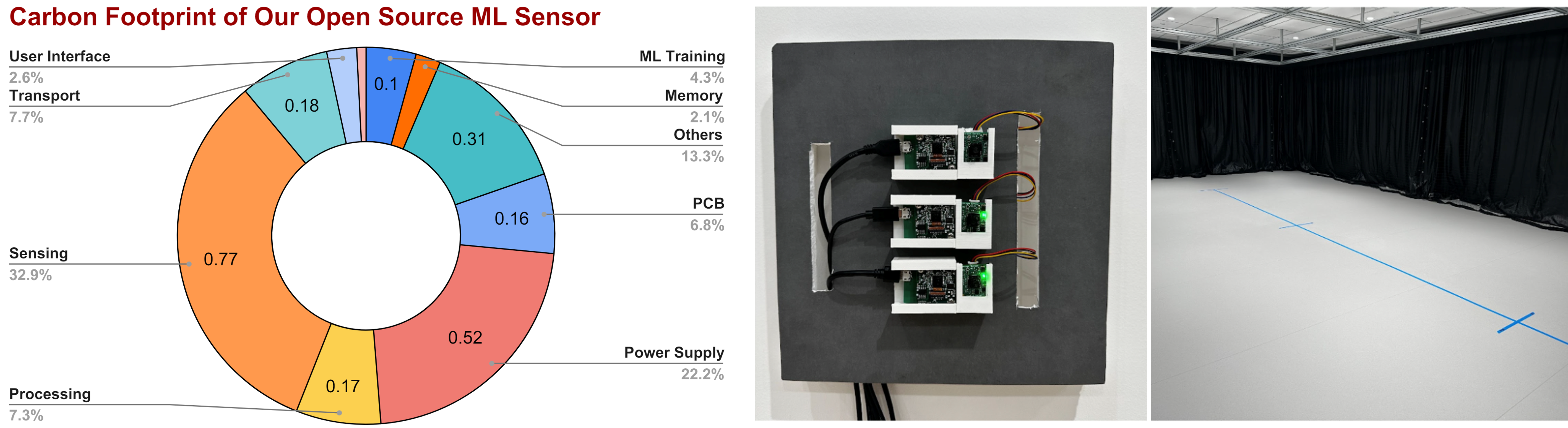}
      \caption{\footnotesize \textit{(Left)} Breakdown by component of the 2.34 kg $\textrm{CO}_2$-eq carbon footprint of our ML sensor, using the TinyML Footprint Calculator~\cite{prakash2023tinyml}. \textit{(Center)} The wall-mounted sensor assembly, consisting of sensors developed in-house on the left and those provided by a commercial partner on the right. \textit{(Right)} The experimental environment with 1m, 3m, and 5m distances marked.}
      \label{fig:env_impact}
\end{figure*}

\textbf{Environmental Footprint.}
We captured the carbon footprint of our ML sensor using the methodology in~\cite{prakash2023tinyml}. The calculator includes fields for processing, sensing, power supply, memory, and more, enabling us to input specifications from our bill of materials. 
Furthermore, we also capture the carbon footprint for the ML sensor’s model training, transport, and three-year use. While training costs can be amortized over multiple sensor deployments, we consider them separately to provide a conservative carbon footprint estimate. The total footprint of our ML sensor, including embodied and operational carbon, is approximately $2.34$ kg $\textrm{CO}_2$-eq. Figure~\ref{fig:env_impact} (left) shows that the majority of the footprint is attributable to the power supply and camera sensor. 
We note that other environmental impact indicators beyond carbon footprint should also be included in future datasheets.

\begin{figure*}[t!]
    \centering
    \includegraphics[width=\textwidth]{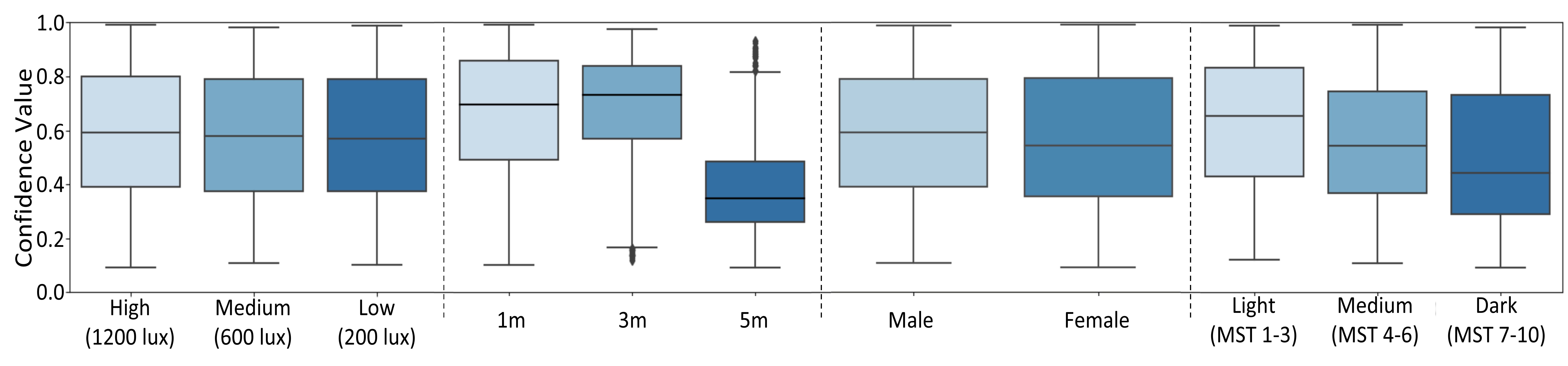}
    \vspace{-4mm}
    \caption{\footnotesize End-to-end performance analysis of the ML sensor tested on 38 volunteers under controlled laboratory conditions. Confidence across lighting conditions \textit{(far left)}, distances \textit{(center left)}, gender \textit{(center right)}, and skin tone \textit{(far right)} estimated using the Monk Skin Tone (MST) Scale \cite{monk_skin_tone}.}
    \label{fig:end}
\end{figure*}

\textbf{End-to-End Performance Analysis:} We present an example of end-to-end performance analysis on our open-source person detection sensor in Figure~\ref{fig:end}. Such an example study was deemed necessary to assess sensor performance in a deployment environment to determine the extent of dataset shift resulting from the use of different hardware (i.e., the onboard camera), embedded demographic biases, as well as biases from environmental changes (e.g., lighting and distance from the camera).

The study room measured 25 x 31 x 10 ft and contained 32 ceiling lights that were uniformly distributed in a 4 x 8 grid. The lighting conditions were captured quantitatively for each participant using a Lux LCD Illuminance Meter (Precision Vision, Inc.) and a C-800-U Spectrometer (Sekonic Corporation).
Sensors were mounted on a wooden board affixed to the wall at a height of 1.5 m above the ground. 39 participants were evaluated at three different distances (1 m, 3 m, and 5 m marked with colored tape) under three lighting conditions (208±31, 584±51, and 1149±59 lux controlled by a dimmer switch). The ambient lighting in the room was provided by artificial lights, and blackout curtains were used to block the ambient lighting from outside (\autoref{fig:env_impact} center and right).
When all the lights were turned on at full strength, the sensor gave an average reading of 1149 lux. The color temperature of the lighting was measured to be 5600 K. 

Participants were asked to provide their gender identity and evaluate their skin tone according to the Monk Skin Tone (MST) Scale. The study evaluated algorithmic bias by bucketing skin tone into three categories: light (MST 0-4), medium (MST 5-7), and dark (MST 8-10). Ten readings from each sensor were averaged at each location and lighting condition.
This anonymous study was approved by the Institutional Review Board of Harvard University on 6 April 2023 (Project Code: IRB23-0136).

The accuracy of the model (see \autoref{fig:end}) is provided as a function of lighting condition, distance, gender identity, and skin tone. We note that overall, 63.2\% of the participants were male, and 36.8\% were female; the percentage of participants corresponding to each skin tone group was: 47.4\% light, 39.4\% medium, and 13.2\% dark.
These analysis provide examples of both device efficacy under changing environmental conditions, a common type of analysis on standard sensor datasheets, as well as possible demographic biases within the model (see Figure~\ref{fig:end}). 

Our performance analysis revealed that while lighting conditions had minimal impact, the model's accuracy degraded sharply when the distance between the subject and the sensor increased from 3 to 5 meters. The model also exhibited potential demographic biases, performing slightly better on men than women and favoring lighter skin tones over darker ones. The diversity of clothing worn by study participants was not fully captured in our testing data, potentially affecting the results. In contrast, the commercial sensor showed increased overall performance and decreased bias, likely due to its more robust custom dataset and advanced software architecture. These findings highlight the importance of considering the sensor's effective range, ensuring diversity and inclusion in training data, actively monitoring and mitigating biases, and investing in high-quality datasets and sophisticated model architectures to enhance accuracy and fairness in ML sensor systems.
\section{Discussion and Limitations} \label{sec:discussion}

Our datasheet template, as proposed, consolidates multiple areas of critical sensor information and finds relevance in numerous practical applications, including predictive maintenance in industrial settings \cite{njor2022primer}, environmental monitoring \cite{gkogkidis2022tinyml}, healthcare diagnostics \cite{tsoukas2021review}, autonomous vehicles \cite{de2021robustifying}, and smart homes \cite{zacharia2022intelligent}. By detailing the hardware characteristics and conformity with industry and regulatory standards, our datasheet provides developers and users with a dependable tool to assess sensor suitability for their specific use-cases. In this section, we discuss the generalizability and limitations of our approach.
Overall, we find that our high-level template can be easily adapted for a wide range of current and future applications but additional development is needed to specify detailed metrics and domain-specific requirements.

\textbf{Open-Source vs. Commercial Comparison.}
At a high-level, the datasheet template was found to be applicable for both our open-source sensor, as well as the commercial sensor, with changes only necessary in a limited number of sections. Sections where changes were necessary were mainly in the data nutrition label and the model characteristics in order to obfuscate aspects of the commercial partner's intellectual property, such as proprietary datasets, models, and training procedures. This obfuscation was critical to enable industrial collaboration and care will need to be taken in the future to ensure that the level of obfuscation balances transparency and intellectual property.

\textbf{Extendability to Varied Data Modalities.}
The structure of the datasheet remains consistent across different modalities, ensuring a familiar framework for assessing the diverse aspects of ML sensors. For instance, when applied to event cameras used in VR/AR~\cite{angelopoulos2020event,gallego2020event}, while event-based cameras have different properties than CMOS cameras and utilize alternative approaches such as spiking networks over convolutional networks, datasheets for sensors using either camera type will retain similar sections such as optical properties of the camera and the network training process.
Similarly, when applied to audio data, instead of detailing the optical properties of the camera, the datasheet would detail the acoustic properties relevant to the microphone, such as sensitivity ranges, signal-to-noise ratios, and the types of audio processing algorithms used.
This adaptability is also true for various model architectures ranging from the CNNs described in our case study to those implementing more basic neural network operations~\cite{ma2023}. In all cases, we will still need to document both the dataset, the resulting metrics for the trained model, and end-to-end metrics.
As such, while incremental refinement might be needed over time, we believe that our datasheet template, regardless of sensor or model configuration can retain the same structure and sections, and simply adjust the details to support the relevant metrics for the specific device.

\textbf{Reliance on Self-Reporting.}
A key limitation of our approach is that the datasheet relies on the accuracy and honesty of the information provided by the manufacturers or developers, with the potential risk of misinformation, misinterpretation, or lack of updates to the datasheet after product updates. Transparency for audit is most useful within relationships of accountability~\cite{cooper2022accountability}, and, as mentioned in Section~\ref{sec:regulation}, oversight mechanisms such as certification from a trusted third-party entity could resolve this concern, and the use of blockchain technologies could aid in auditability~\cite{ali2021comparative}.

\textbf{Cross-Compatibility of Metrics.}
Our case study provides an important example of another key limitation of our datasheet which is cross-compatibility and standardization of metrics across a wide possible breadth of ML Sensor device designs and implementations. In fact, we found it challenging to directly compare end-to-end results from the commercial and open-source devices due to the differing approaches utilized by the sensors. The commercial sensor utilized a face detection bounding-box model with a detection threshold set at $\sim$0.6, whereas our open-source sensor focused on person detection within the full image. This along with differences in camera specification meant that the open-source device was better at detecting individuals over longer distances, while the commercial sensor had a wider angle of detection. This suggests that future research is needed to design and build methods to fairly compare and evaluate ML sensors as their diversity grows. However, we believe that ensuring that devices are bundled with relevant, even if not cross-compatible, end-to-end metrics is a critical first step to ensuring transparency and auditability for these devices.
\section{Conclusion and Future Work} \label{sec:conclusion}
This paper introduces a novel datasheet template designed specifically for ML sensors to provide meaningful ML sensor transparency. We focus on embedded ML deployments, where models are tightly integrated and fused with devices, warranting closer introspection in this new space. We demonstrate the practical application of our datasheets by developing them for two real-world ML sensors, enhancing transparency, auditability, and user-friendliness across open-source and commercial devices. By providing a standardized format for documenting and evaluating ML sensors, our work contributes to the advancement of data-centric ML research and the creation of reliable benchmarks for next-generation ML that is deeply fused in with embedded and edge AI systems.

In future work we hope to expand upon our initial case study and provide formal documentation to help others build their own datasheets for their own ML Sensors. We also intend to perform large-scale validation of our datasheet template design, approach, layout, and content to ensure its effectiveness and generality. In particular, large-scale user studies across academia and industry can help analyze the effectiveness of the current template. To that end, future work will also study the effectiveness of the datasheet as a component of audit, to better understand design changes that could promote transparency for auditability across stakeholder groups. Similarly, large-scale design studies to understand what forms of visualizations, and organization of the data contained in these datasheets would make them most effective in conveying their information would further improve the impact of this work. Finally, we hope to find ways to build on recent work (e.g., Croissant-RAI~\cite{jain2024standardized}) to ensure that future datasheets are both human and machine readable to increase their impact and usability for a wider range of users and applications.

\bibliographystyle{ieeetr}
\bibliography{refs}

\appendices
\includepdf[pages=-]{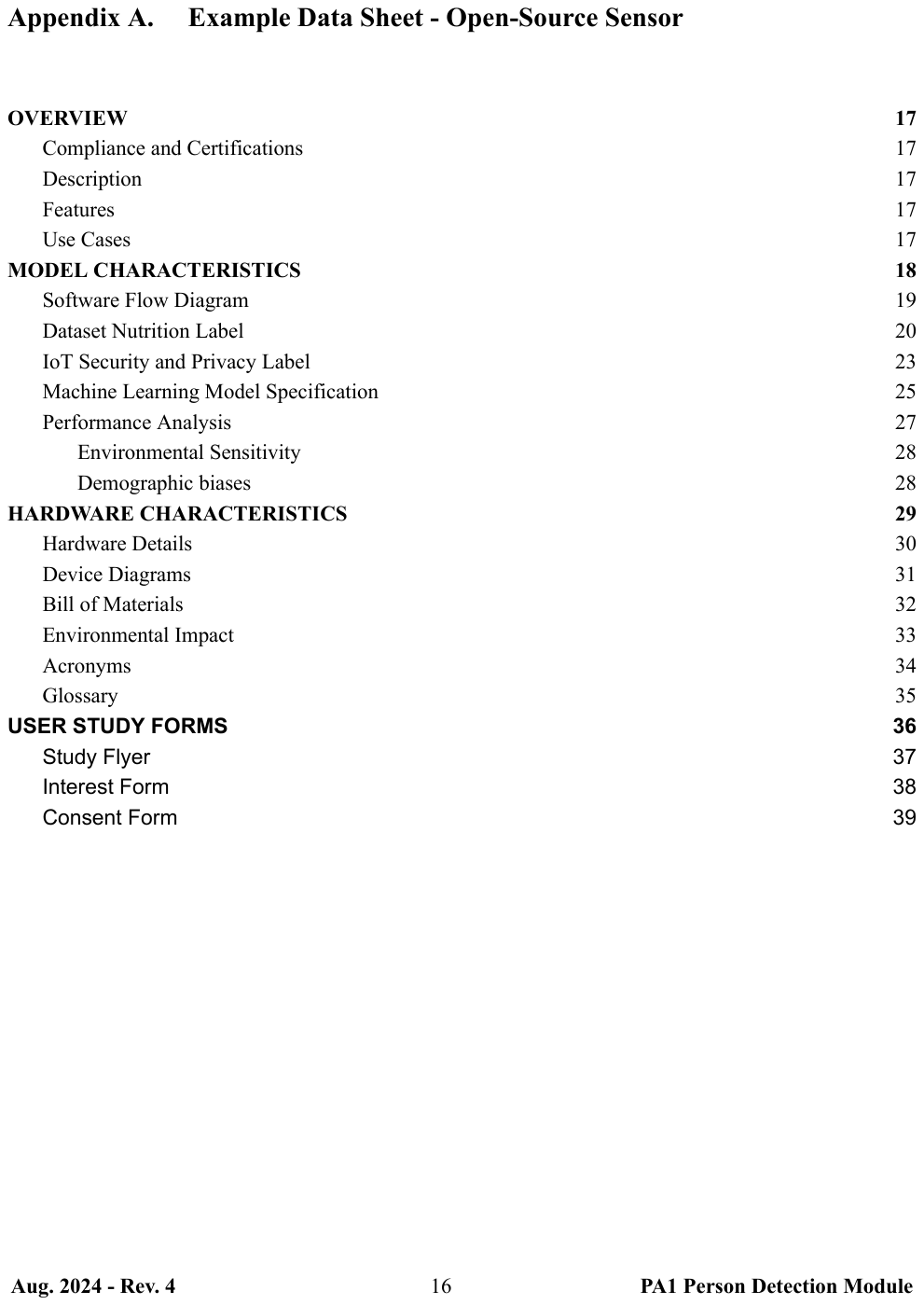}
\includepdf[pages=-]{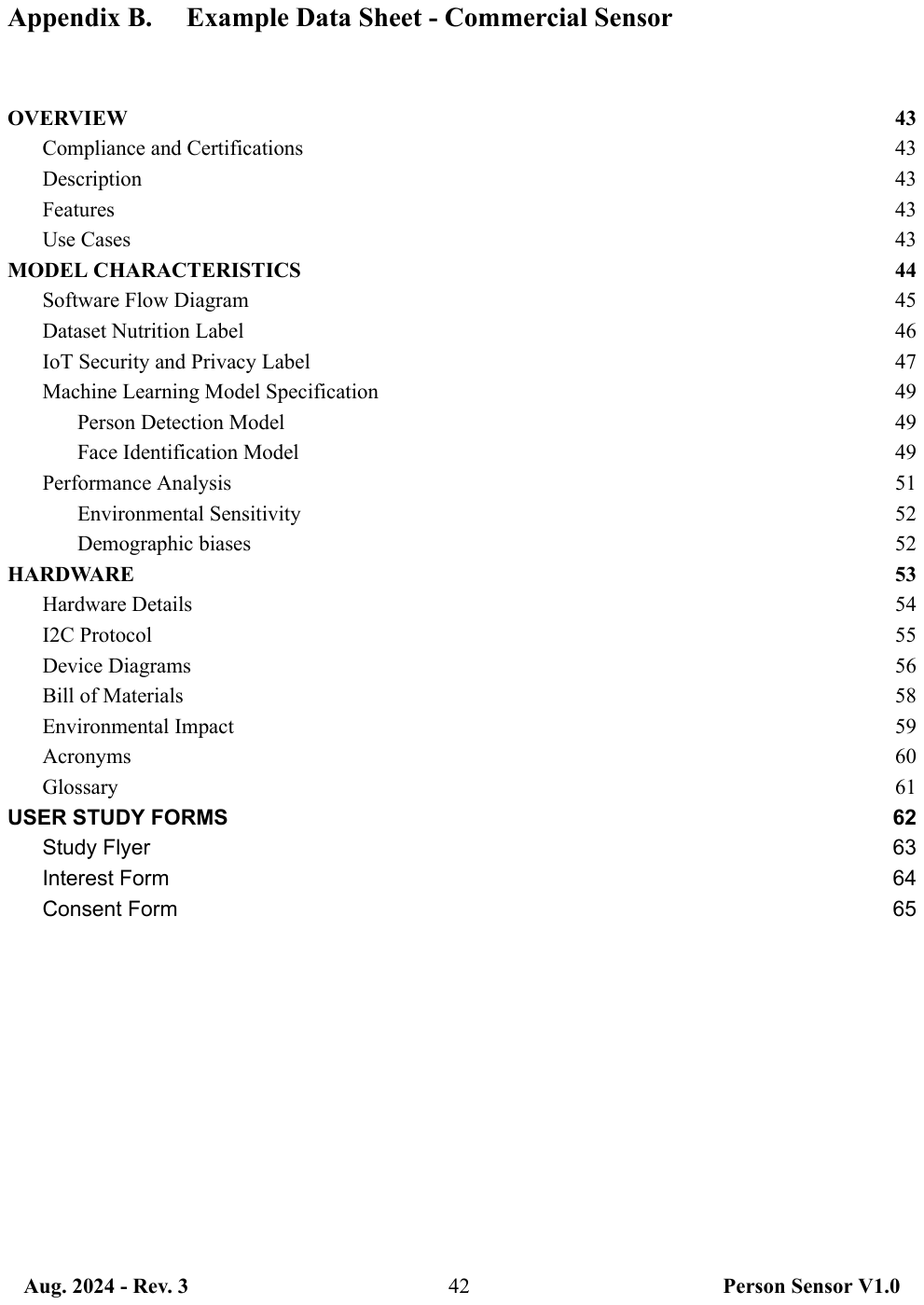}

\end{document}